\documentclass[letterpaper]{article} 
\usepackage{aaai2026}  
\usepackage{times}  
\usepackage{helvet}  
\usepackage{courier}  
\usepackage[hyphens]{url}  
\usepackage{graphicx} 
\usepackage{natbib}  
\usepackage{caption} 
\usepackage{algorithm}
\usepackage{algorithmic}
\usepackage{amssymb}
\usepackage{amsmath}
\usepackage{booktabs}
\usepackage{makecell}
\usepackage{cleveref}
\usepackage{dsfont}
\usepackage{enumitem}
\usepackage{multirow}
\usepackage{xcolor}
\definecolor{sblue}{rgb}{0.9, 0.9, 1.0}
\definecolor{instructbg}{gray}{0.94}

\usepackage{colortbl}
\usepackage{caption}
\usepackage{subcaption}
\usepackage{pifont}
\newcommand{\xmark}{\ding{55}}%
\usepackage{float}
\newcommand{\pub}[1]{\tiny\textcolor{deemph}{[#1]}}
\definecolor{deemph}{gray}{0.4}

\usepackage{newfloat}
\usepackage{listings}
\DeclareCaptionStyle{ruled}{labelfont=normalfont,labelsep=colon,strut=off} 
\floatstyle{ruled}
\newfloat{listing}{tb}{lst}{}
\floatname{listing}{Listing}
\title{SkyMoE: A Vision-Language Foundation Model for Enhancing Geospatial Interpretation with Mixture of Experts}
\author{
    Jiaqi Liu\textsuperscript{\rm 1,\rm 2}, Ronghao Fu\textsuperscript{\rm 1,\rm 2}\thanks{Corresponding authors.}, Lang Sun\textsuperscript{\rm 1,\rm 2}, Haoran Liu\textsuperscript{\rm 1,\rm 2}, Xiao Yang\textsuperscript{\rm 1,\rm 2}, Weipeng Zhang\textsuperscript{\rm 1,\rm 2}, \\ Xu Na\textsuperscript{\rm 1,\rm 2}, Zhuoran Duan\textsuperscript{\rm 1,\rm 2}, Bo Yang\textsuperscript{\rm 1,\rm 2}\footnotemark[1]
}
\affiliations{
    \textsuperscript{\rm 1}College of Computer Science and Technology, Jilin University, China\\
    \textsuperscript{\rm 2}Key Laboratory of Symbolic Computation and Knowledge Engineering of Ministry of Education, Jilin University, China\\

    \{liujq21,sunlang24,lhr24,yangx23,zhangwp24,naxu23,duanzr24\}@mails.jlu.edu.cn, \{furh,ybo\}@jlu.edu.cn 
}

\usepackage{bibentry}

\begin{document}

\maketitle

\begin{abstract}
The emergence of large vision-language models (VLMs) has significantly enhanced the efficiency and flexibility of geospatial interpretation. However, general-purpose VLMs remain suboptimal for remote sensing (RS) tasks. Existing geospatial VLMs typically adopt a unified modeling strategy and struggle to differentiate between task types and interpretation granularities, limiting their ability to balance local detail perception and global contextual understanding. In this paper, we present SkyMoE, a Mixture-of-Experts (MoE) vision-language model tailored for multimodal, multi-task RS interpretation. SkyMoE employs an adaptive router that generates task- and granularity-aware routing instructions, enabling specialized large language model experts to handle diverse sub-tasks. To further promote expert decoupling and granularity sensitivity, we introduce a context-disentangled augmentation strategy that creates contrastive pairs between local and global features, guiding experts toward level-specific representation learning. We also construct MGRS-Bench, a comprehensive benchmark covering multiple RS interpretation tasks and granularity levels, to evaluate generalization in complex scenarios. Extensive experiments on 21 public datasets demonstrate that SkyMoE achieves state-of-the-art performance across tasks, validating its adaptability, scalability, and superior multi-granularity understanding in remote sensing.
\end{abstract}


\section{Introduction}
Recent advances in artificial intelligence have led to the rapid development of Vision-Language Models (VLMs), which demonstrate impressive generalization across a wide range of visual and linguistic tasks ~\cite{wang2025deep,liu2024enhancing,durante2024agent}. Their success is largely attributed to the ability to extract rich, high-level semantic representations from images, enabling capabilities such as open-vocabulary recognition, implicit reasoning, and multimodal alignment. These strengths have sparked growing interest in applying VLMs to Remote Sensing (RS), where tasks such as land cover classification and open-set object detection demand sophisticated scene understanding ~\cite{wang2023open,zhang2024earthgpt}. 

\begin{figure}[!t]
    \centering
    \includegraphics[width=\linewidth]{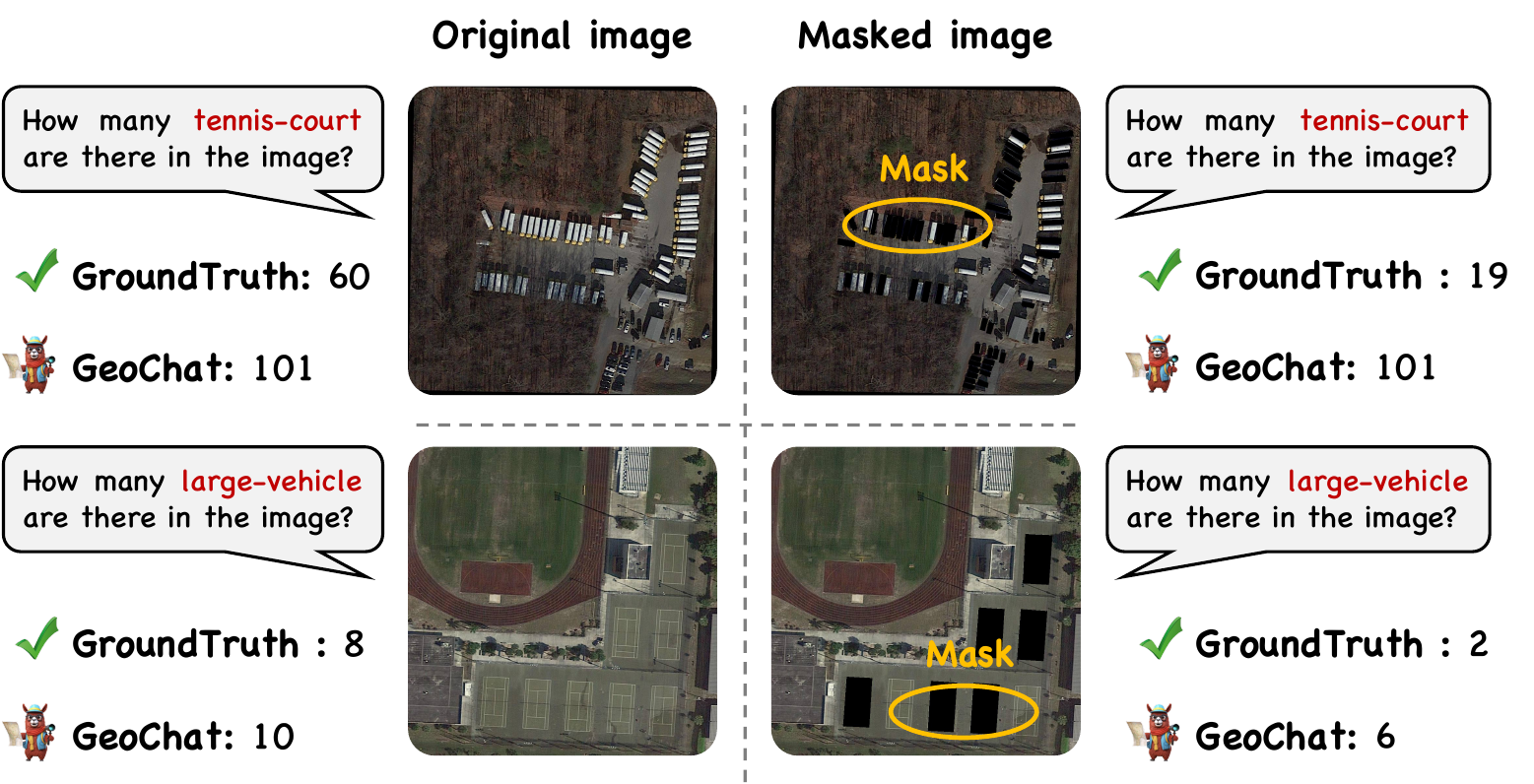}
    \caption{Selective masking of objects in images reveals minimal variation in model-provided counts, indicating a reliance on background context over precise enumeration.}
    \label{fig:maskcomp}
\end{figure}

\begin{figure*}[!t]
   \centering
   \includegraphics[width=\linewidth]{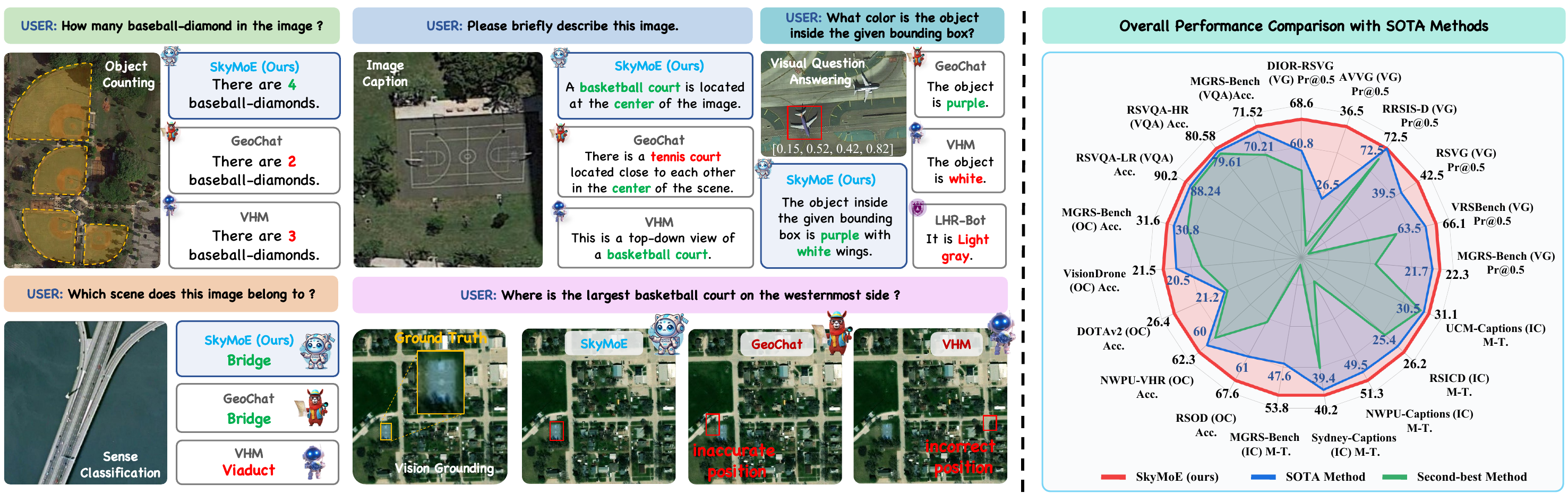}
   \caption{Overall performance comparison between SkyMoE and eight state-of-the-art models across 21 datasets spanning five remote sensing interpretation tasks. The radar chart shows that SkyMoE achieves competitive performance on most benchmarks.}
   \label{fig:main}
\end{figure*}

However, RS imagery presents distinct challenges that set it apart from conventional vision tasks. Specifically, it requires a dual capacity to capture global spatial layouts over large geographic extents while simultaneously recognizing fine-grained, localized objects. This coarse-to-fine semantic demand often exceeds the representational limits of standard VLMs, which typically rely on monolithic architectures with shared parameters. Such models struggle to reconcile the competing needs for detailed object-level recognition and holistic scene-level reasoning.

This inherent tension is evident in recent remote sensing vision language models (RS-VLMs). As illustrated in~\Cref{fig:maskcomp}, masking-based analyses reveal a strong overreliance on global context. Even when key foreground entities (e.g., vehicles, courts) are occluded, models often produce inflated object counts—indicating a lack of instance-level discrimination and an overdependence on background priors. This behavior highlights a critical shortcoming: the inability to disentangle and properly integrate localized and contextual features, which undermines both model reliability and task-specific performance.

To address this issue, there is increasing interest in modular architectures that support specialization, such as Mixture-of-Experts (MoE) frameworks. MoE architectures offer structural flexibility by distributing computation across expert subnetworks. However, prior RS-based MoE models~\cite{lin2025rs} have mostly adopted general-purpose training paradigms without mechanisms to enforce meaningful expert specialization. As a result, experts often learn overlapping functions, failing to leverage the MoE design for effective feature disentanglement.

In this work, we introduce SkyMoE, a novel VLM framework tailored for multi-scale RS interpretation. SkyMoE integrates two co-designed components: (1) an MoE-based architecture that enables modular specialization, and (2) a contrastive data augmentation strategy that introduces explicit inductive biases to steer expert learning. By systematically modifying local object attributes while preserving global context, we generate fine-grained training samples that compel each expert to focus on either localized or global semantics. This targeted supervision facilitates effective task decomposition and unlocks the MoE architecture’s potential for interpretable and robust representation learning. To facilitate comprehensive evaluation, we additionally introduce MGRS-Bench, a dedicated benchmark encompassing a broad range of RS tasks at varying semantic granularities. As shown in ~\Cref{fig:main}, the proposed SkyMoE exhibits competitive performance across multiple tasks. In summary, this work has the following contributions:

\begin{itemize}
    \item We propose SkyMoE, a novel MoE-based architecture that dynamically routes remote sensing tasks to specialized experts, guided by task-specific interpretation granularity and semantic complexity.
    
    \item We design a context-disentangled data augmentation strategy that introduces inductive biases to facilitate expert specialization within the MoE framework.
    
    \item We construct MGRS-Bench, a comprehensive benchmark covering diverse tasks, granularity levels, and resolution variations for RS-VLM evaluation.

    \item Through extensive experiments on 21 diverse benchmarks, we demonstrate that SkyMoE establishes a new state-of-the-art, achieving superior results across a wide spectrum of both fine-grained and scene-level tasks.
\end{itemize}

\section{Related Work}

\subsection{Vision-Language Models in General Domain}

The integration of visual encoders with powerful, open-source Large Language Models (LLMs) has given rise to a new paradigm of Vision-Language Models (VLMs) that have demonstrated remarkable cross-domain capabilities. Models such as LLaVA~\cite{liu2023visual}, MiniGPT-4 ~\cite{zhu2023minigpt}, and Qwen-VL ~\cite{baiqwen} typically employ a unified module to project encoded visual features into the LLM's embedding space. This architectural design, while successful for general-domain tasks (e.g., image captioning, visual dialogue) that prioritize holistic scene understanding, inherently encourages the learning of global contextual features at the expense of fine-grained local details. Consequently, when confronted with specialized domains like remote sensing, which are characterized by dense small objects and demand high-fidelity localization, these general-purpose VLMs often exhibit suboptimal performance. Even advanced models like DeepSeek-VL~\cite{wudeepseek}, which incorporates an MoE framework, primarily leverage it for scaling model capacity on general tasks, rather than explicitly addressing the fundamental tension between local and global feature representation. This highlights the need for a VLM framework specifically architected and trained for the unique challenges of remote sensing imagery.

\subsection{Vision-Language Models in Remote Sensing}

To address domain-specific challenges, the remote sensing community has developed specialized VLMs. Models like GeoChat~\cite{kuckreja2024geochat} and EarthGPT~\cite{zhang2024earthgpt} have been tailored for geospatial tasks, demonstrating the potential of VLMs in this field. However, these models still largely inherit the architectural tendency of their general-domain counterparts to prioritize global context, thus struggling to achieve a robust balance with local details. Notably, some works have attempted to explicitly address this imbalance. For instance, VHM~\cite{pang2025vhm} mitigates this issue to an extent by fusing multi-level features from its vision encoder. However, this represents a static, task-agnostic fusion strategy. The combination of features is pre-defined by the architecture and does not adapt to the specific demands of a given input, which may require a dynamic emphasis on either local or global information. This highlights that while the problem of feature balancing is recognized, a mechanism for dynamic, input-aware allocation of model resources remains a critical unfilled gap in RS-VLMs.

\subsection{Mixture of Experts for Remote Sensing}
The promise of MoE as a dynamic, learned framework has recently led to its adoption in the remote sensing domain. For example, RS-MoE~\cite{lin2025rs} employs an MoE structure to effectively fuse information from heterogeneous data sources, while RSUniVLM~\cite{liu2024rsunivlm} utilizes experts to specialize in different downstream tasks. These pioneering works demonstrate the versatility of the MoE paradigm. However, in these existing frameworks, the MoE mechanism is primarily leveraged for multi-modal fusion or task-level specialization, rather than being explicitly directed at resolving the fundamental, underlying tension between local and global feature representations. Consequently, while they utilize an MoE architecture, their training paradigms are not designed to compel expert differentiation for the specific purpose of feature balancing. As such, the full potential of MoE as a solution to this core challenge remains latent and unrealized, underscoring the novelty and necessity of our proposed approach.

\begin{figure*}[!t]
    \centering
    \includegraphics[width=\linewidth]{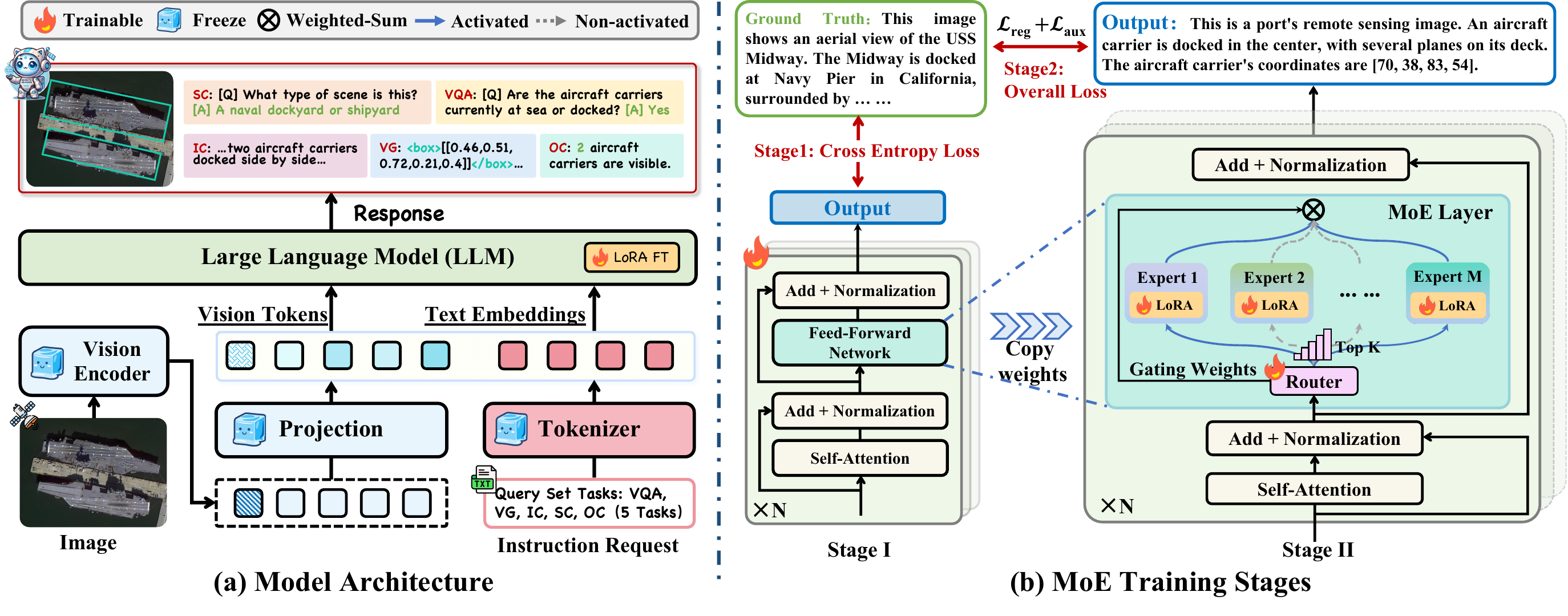}
    \caption{\textbf{Training framework and strategy.} SkyMoE adopts the standard vision-language framework composed of an image encoder, visual adaptor, and decoder-only LLM. The training employs a two-phase approach, Stage \uppercase\expandafter{\romannumeral1}: Initial LLM pretraining establishes multimodal understanding without MoE layers, followed by Stage \uppercase\expandafter{\romannumeral2}: MoE specialization through expert initialization via cloned FFN weights and subsequent fine-tuning.}
    \label{fig:framework}
\end{figure*}

\section{Methodology}
The architecture of SkyMoE, as depicted in~\Cref{fig:framework}. We begin by describing the architecture in detail. Then, we present the training methodology specifically designed for SkyMoE. Finally, we discuss the training objectives that guide its optimization.

\subsection{Architecture of SkyMoE}
\label{sec:model}
Given an RGB image \(\mathbf{v} \in \mathbb{R}^{H \times W \times 3}\), where \(H\) and \(W\) denote the original resolution, we utilize the pretrained vision backbone of CLIP-ViT(L-14)~\cite{tay2017learning}, which segments the image into 576 patches at an input resolution of \(336 \times 336\). To enhance the model's ability to capture intricate details and small objects in remote sensing imagery, we interpolate the positional encoding in the transformer-based CLIP model~\cite{tay2017learning} to support input image sizes of \(504 \times 504\). This enhancement effectively doubles the number of patches to 1296 per image, significantly improving visual grounding in high-resolution remote sensing images.

The vision encoder processes the input image to generate a visual token sequence \(\mathcal{Z} = [z_{1}, z_{2}, \ldots, z_{P}] \in \mathbb{R}^{P \times C}\), with \(P = \frac{H \times W}{36^2}\) representing the sequence length of visual tokens. A visual projection layer \(f\) maps this sequence from \(\mathbb{R}^{P \times C}\) to \(\mathbb{R}^{P \times D}\), aligning the visual tokens with the hidden size \(D\) of the large language model. Concurrently, the text component undergoes a word embedding layer \(g\), resulting in sequence tokens \(\mathcal{T} = [t_{1}, t_{2}, \ldots, t_{N}] \in \mathbb{R}^{N \times D}\), where \(N\) signifies the sequence length of text tokens.

The visual tokens \(\mathcal{V}\) and text tokens \(\mathcal{T}\) are concatenated to form a unified sequence \(\mathbf{x}_0 = [v_{1}, v_{2}, \ldots, v_{P}, t_{1}, t_{2}, \ldots, t_{N}]\). This combined sequence is then fed into the LLM, which comprises stacked multi-head self-attention (MSA) and feed-forward neural networks (FFN) blocks. Each block incorporates layer normalization (LN) and residual connections, enhancing the model's capacity for complex pattern recognition and feature extraction. The forward pass through the LLM can be mathematically represented as follows:
\begin{equation}
\small
    \begin{aligned}
        \mathbf{x}_{\ell}^{\prime} &= \mathrm{MSA}(\mathrm{LN}(\mathbf{x}_{\ell-1})) + \mathbf{x}_{\ell-1}, \quad \ell = 1, \ldots, L, \\
        \mathbf{x}_{\ell} &= \mathrm{MoE}(\mathrm{LN}(\mathbf{x}_{\ell}^{\prime})) + \mathbf{x}_{\ell}^{\prime}, \quad \ell = 1, \ldots, L,
    \end{aligned}
\end{equation}
where \(L\) denotes the number of layers in the LLM. The final output \(\mathcal{Y}\) is obtained after the last layer through layer normalization:
\begin{equation}
\small
    \mathcal{Y} = \mathrm{LN}(\mathbf{x}_L).
\end{equation}

The MoE layer, a pivotal component of our model, enhances the LLM's adaptability by incorporating multiple FFNs. Initially, we replicate the FFNs from stage \uppercase\expandafter{\romannumeral2} to form an ensemble of experts \(\mathcal{E} = [e_{1}, e_{2}, \ldots, e_{E}]\). A router, implemented as a linear layer, predicts the probability of each token being assigned to each expert. This probability distribution is formulated as:
\begin{equation}
\small
    \mathcal{P}(\mathbf{x})_i = \frac{e^{f(\mathbf{x})_i}}{\sum_{j=1}^E e^{f(\mathbf{x})_j}},
\end{equation}
where the router produces weight logits \(f(\mathbf{x}) = \mathbf{W} \cdot \mathbf{x}\), normalized by the softmax function. Here, \(\mathbf{W} \in \mathbb{R}^{D \times E}\) represents the lightweight training parameters, and \(E\) denotes the number of experts. Each token is processed by the top-\(k\) experts with the highest probabilities, and the final output is a weighted sum based on these probabilities:
\begin{equation}
\small
    \mathrm{MoE}(\mathbf{x}) = \sum_{i=1}^k \mathcal{P}(\mathbf{x})_i \cdot \mathcal{E}(\mathbf{x})_i.
\end{equation}

\subsection{Dual-Stage Training}
\label{sec:train}
We employ a two-stage strategy for training SkyMoE. Utilizing 6 NVIDIA A800-80G GPUs, we train the model with a batch size of 144 for 5 epochs. The AdamW optimizer~\cite{loshchilov2017decoupled} is used with an initial learning rate of 2e-5, combined with a cosine learning rate scheduling strategy. All components of SkyMoE are optimized with the large-scale instruction dataset, the construction of which is introduced in the \textit{Datasets and Evaluation Benchmarks} section, to incorporate RS visual knowledge into the model.

\textbf{Stage \uppercase\expandafter{\romannumeral1}:} To enhance the effectiveness of our model on general visual tasks and optimize training efficiency, we employ a strategy that involves initializing the network with pre-trained weights and fine-tuning specific segments for remote sensing related tasks. Specifically, we utilize a pre-trained CLIP-ViT(L-14) encoder~\cite{tay2017learning}, trained on large amounts of textual and visual data, a pre-trained MLP adaptor~\cite{liu2024improved}, trained on a 558K subset of the LAION-CC-SBU dataset~\cite{schuhmann2022laion} with BLIP~\cite{li2022blip} captions, and Vicuna-v1.5~\cite{shen2023hugginggpt} to initialize our model. To adapt our model to remote sensing images, we subsequently apply LoRA~\cite{lora} fine-tune on the LLM, targeting the \(W_q\) and \(W_v\) matrices with a designated rank \(r\) set to 64, while keeping the MLP adaptor and the CLIP encoder~\cite{tay2017learning} frozen during training. The model undergoes training consistently at an image resolution of 504 $\times$ 504 throughout the whole process. This stage not only lays the foundation for the model's ability to handle general visual tasks but also enhances its capabilities and controllability by tuning it with multi-modal instruction data~\cite{kuckreja2024geochat}. We use more complex instructions, including tasks such as image logical reasoning and text recognition, which require the model to have a stronger multi-modal understanding. Typically, for dense models, the VLM training is considered complete at this stage. However, we encounter challenges in simultaneously transforming the LLM into an VLM and sparsifying the VLM. To address this, SkyMoE utilizes the weights from this stage as initialization for the next stage to alleviate the learning difficulty of the sparse model.

\textbf{Stage \uppercase\expandafter{\romannumeral2}:} In the second stage, we transform the model into a sparse VLM by integrating a MoE architecture. Expert modules are initialized by replicating the FFN multiple times, and MoE layers are interleaved with standard MLP layers. Specifically, every second layer is replaced with an MoE layer. When image tokens and text tokens are fed into the MoE layers, 
the router calculates the matching weights between each token and the experts. Each token is then processed by the top-2 experts, and the outputs are aggregated by weighted summation based on the router's weights. Non-selected experts remain inactive, enabling dynamic sparsity and efficient computation. Expert parallel size is set to 1 to support distributed expert execution. This architecture enables SkyMoE to adaptively select computation paths, achieving both high-level reasoning and fine-grained perceptual understanding across diverse remote sensing tasks.

\subsection{Training Objectives}
\label{sec:objective}

The total loss function \(\mathcal{L}_{\text{total}}\) is composed of the auto-regressive loss \(\mathcal{L}_{\text{regressive}}\) and the auxiliary loss \(\mathcal{L}_{\text{aux}}\), with the auxiliary loss being scaled by a balancing coefficient \(\alpha\):
\begin{equation}
\small
  \mathcal{L}_{\text{total}} = \mathcal{L}_{\text{regressive}} + \alpha \cdot \mathcal{L}_{\text{aux}}.
  \label{eq:total_loss}
\end{equation}

\textbf{Auto-Regressive Loss.} The output of the LLM is optimized via a generative loss in an auto-regressive fashion. Given an image and text, SkyMoE produces the output sequence \(\mathcal{Y} = [y_{1}, y_{2}, \ldots, y_{K}] \in \mathbb{R}^{K \times D}\) by sequentially generating each element, where \(K = P + N\) denotes the length of the output sequence. The loss function is defined as:
\begin{equation}
\small
  \mathcal{L}_{\text{regressive}} = -\sum_{i=1}^N \log \ p_\theta\left(\mathcal{Y}^{[P+i]} \mid \mathcal{V}, \mathcal{T}^{[:i-1]}\right),
  \label{eq:regressive_loss}
\end{equation}
where \(\theta\) represents the trainable parameters, and the loss is computed solely for the generated text.

\textbf{Auxiliary Loss.} Given the presence of multiple experts, it is essential to enforce load balancing constraints on the MoE layer. We integrate a differentiable load balancing loss into each MoE layer to promote balanced token handling among experts:
\begin{equation}
    \mathcal{L}_{\text{aux}} = E \cdot \sum_{i=1}^{E} \mathcal{F}_i \cdot \mathcal{G}_i,
\end{equation}
where \(\mathcal{F}\) denotes the fraction of tokens processed by each expert \(\mathcal{E}_i\), and \(\mathcal{G}\) represents the average routing probability for \(\mathcal{E}_i\), which are given by:
\begin{equation}
\small
\mathcal{F} = \frac{1}{K} \sum_{i=1}^E \mathds{1}\{\operatorname{argmax} \mathcal{P}(\mathbf{x}) = i\}, 
\mathcal{G} = \frac{1}{K} \sum_{i=1}^{K} \mathcal{P}(\mathbf{x})_i.     
\end{equation}
\section{Datasets and Evaluation Benchmarks}
\label{sec:dataset}
\begin{figure}[!t]
    \centering
    \includegraphics[width=\linewidth]{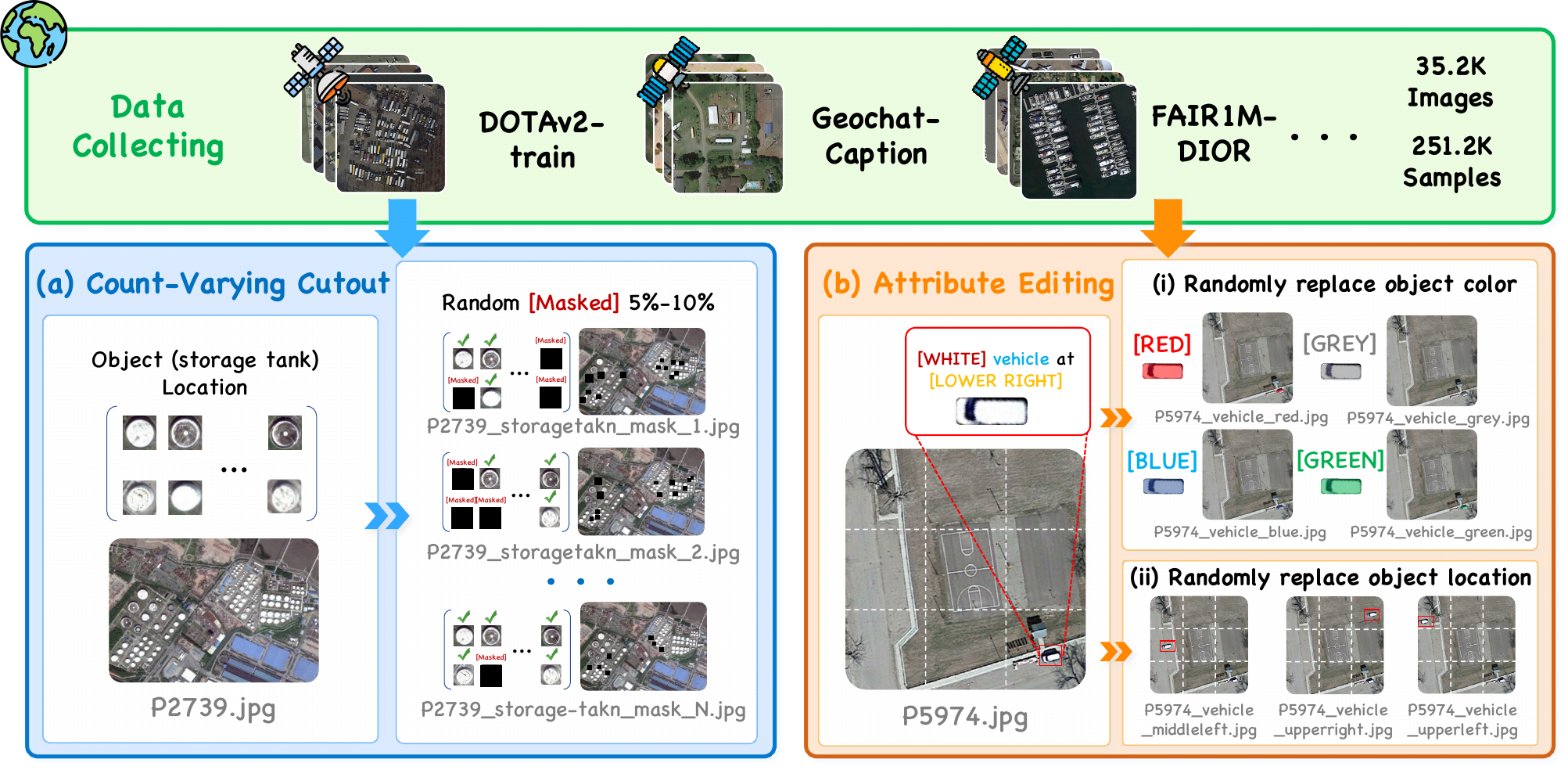}
    \caption{Context-Disentangled Data Augmentation.}
    \label{fig:aug}
\end{figure}

To support the training and evaluation of SkyMoE, we construct a comprehensive dataset pipeline encompassing both foundational data aggregation and targeted augmentation. In Stage~\uppercase\expandafter{\romannumeral1}, we aggregate diverse RS datasets spanning multiple tasks to form a unified instruction-following training corpus. In Stage~\uppercase\expandafter{\romannumeral2}, we apply task-specific augmentation strategies (Figure~\ref{fig:aug}) to enhance data diversity and optimize the MoE fine-tuning process. Together, these stages ensure that SkyMoE is exposed to a wide range of task types, object densities, and attribute variations—ultimately boosting its generalization ability in real-world RS scenarios. In addition, we introduce MGRS-Bench, a purpose-built benchmark that enables multi-granularity and cross-resolution evaluation, thereby filling critical gaps in existing test sets.

\subsection{Instruction Dataset Construction}

To enable SkyMoE to handle diverse RS tasks in a unified manner, we construct a multi-task instruction dataset by curating and integrating multiple high-quality, publicly available sources. The dataset spans five vision-language tasks. Our final training set includes over 251k instruction samples, with an additional 11k instances for testing. The training data is derived from established benchmarks such as DOTA~\cite{xia2018dota}, DIOR~\cite{li2020object}, FAIR1M~\cite{sun2022fair1m} , DOTA-v2~\cite{xia2018dota}. To ensure robust generalization, the test set is sourced independently from datasets like DIOR-RSVG~\cite{zhan2023rsvg}, RSOD~\cite{long2017accurate}, and NWPU-RESISC45-test~\cite{cheng2017remote}. A full list of source datasets is available in the supplementary material.

To unify training across tasks, we reformat each sample into an instruction-following format. For this, we design task-specific templates that convert original annotations into natural language instructions and standardized answers. For example, scene classification is expressed as \textit{``Which scene does this image belong to?''}, expecting a category name as output. Visual grounding uses the prompt \textit{``[refer] Where is \texttt{<}p\texttt{>} referring expression \texttt{<}/p\texttt{>}?''}, requiring bounding box coordinates. Similar templates are used for counting, captioning, and VQA. This unified structure allows the model to learn diverse tasks under a single conversational paradigm.

\begin{table*}[!tb]
\begin{center}

\resizebox{\textwidth}{!}{
\begin{tabular}{lcccccccccccccccccc} 
\toprule
\multirow{2}{*}{\textbf{Method}} &\multicolumn{3}{c}{\textbf{UCM-Captions}} & \multicolumn{3}{c}{\textbf{RSICD}} & \multicolumn{3}{c}{\textbf{RSITMD}} & \multicolumn{3}{c}{\textbf{NWPU-Captions}} & \multicolumn{3}{c}{\textbf{Sydney-Captions}} & \multicolumn{3}{c}{\textbf{MGRS-Bench}} \\
\cmidrule(lr){2-4} \cmidrule(lr){5-7} \cmidrule(lr){8-10} \cmidrule(lr){11-13} \cmidrule(lr){14-16} \cmidrule(lr){17-19}
 &\textbf{B-4} & \textbf{MT}  & \textbf{R-L}
 &\textbf{B-4} & \textbf{MT}  & \textbf{R-L} 
 &\textbf{B-4} & \textbf{MT}  & \textbf{R-L} 
 &\textbf{B-4} & \textbf{MT}  & \textbf{R-L} 
 &\textbf{B-4} & \textbf{MT}  & \textbf{R-L} 
 &\textbf{B-4} & \textbf{MT}  & \textbf{R-L}  \\

\midrule
\multicolumn{7}{l}{\textcolor{gray}{\textit{Close-source Commercial Vision-Language Models}}} \\ 
GPT-4V\nocite{gpt-4v}  \pub{OpenAI'24} & 28.4 & 25.6 & 17.8 & 16.8 & 16.7 & 15.9 & 27.3 & 21.1 & 14.0 & 39.6 & 25.7 & 20.9 & 28.5 & 24.5 & 17.5 & 22.1 & 32.6 & 22.3     \\ 
\midrule
\multicolumn{19}{l}{\textcolor{gray}{\textit{Open-source Vision-Language Models}}} \\
MiniGPT-v2\nocite{chenminigpt} \pub{Meta'24} & 18.1 & 21.7  & 11.4 & 11.9 & 15.8  & 11.2 & 19.0 & 19.7  & 10.0 & 28.6 & 23.3  & 13.0 & 19.8 & 19.5  & 10.6 & 19.4 & 15.9  & 16.9  \\
LLaVA-1.5\nocite{liu2023llava} \pub{CVPR'24}& 24.6 & 26.7  & 17.1 & 21.3 & 16.4  & 17.9 & 35.0 & 23.1  & 16.7 & 47.6 & 26.5  & 21.9 & 39.1 & 25.3  & 21.1  & \underline{36.6} & 28.2  & 36.0 \\
Qwen2.5-VL\nocite{bai2025qwen2} \pub{BABA'24} & 14.7 & 24.5  & 11.5 & 9.5 & 17.2  & 11.2 & 16.6 & 21.8  & 10.5 & 24.1 & 26.4  & 14.4 & 16.1 & 22.5  & 11.3 &  30.7 & \underline{47.6}  & \textbf{37.8} \\
DeepSeek-VL\nocite{wudeepseek} \pub{DeepSeek'24} & 23.1 & \underline{30.5}  & 16.6 & 13.9 & 20.6  & 14.6 & 26.9 & 25.0  & 14.6 & 39.1 & 30.5  & 20.4 & 38.2 & 26.1  & 20.7 & 31.6 & 28.0  & 33.3 \\
\midrule 
\multicolumn{19}{l}{\textcolor{gray}{\textit{Open-source Remote Sensing Vision-Language Models}}} \\
GeoChat\nocite{kuckreja2024geochat} \pub{CVPR'24}  & 21.0 & 20.8  & 14.1 & 15.8 & 14.1 & 13.4 & 27.5 & 20.3  & 13.8 & 37.8 & 25.2  & 16.3 & 24.6 & 16.5  & 10.5  & 24.4 & 13.9  & 22.7 \\
VHM\nocite{pang2025vhm}  \pub{AAAI'25} & \underline{42.4} & 26.9 &  \textbf{24.9} & 19.5 & 22.2  & 18.0 & 36.4 & 17.9  & 15.7 & 47.9 & 18.9  & 17.8 & 37.0 & 21.5  & 17.4 & 29.1 & 21.9  & 28.3  \\
SkySenseGPT\nocite{luo2024skysensegpt} \pub{arXiv'24} & 15.7 & 23.3  & 10.9 & 12.2 & 17.0  & 12.8 & 20.6 & 21.7 &  11.8 & 28.2 & 24.7  & 14.4 & 18.4 & 21.3  & 11.9 & 24.9 & 17.2  & 25.2 \\
LHRS-Bot\nocite{muhtar2024lhrs} \pub{ECCV'24} & 23.5 & 30.1 &  17.1 & 21.2 & 23.8  & 19.4 & \textbf{38.2} & \underline{27.9} &  \textbf{19.8} & \underline{58.4} & \underline{49.5}  & \underline{35.8} & 40.2 & 36.2 &  25.1 & 25.6 & 20.8  & 25.2 \\
RSUniVLM\nocite{liu2024rsunivlm} \pub{arXiv'24} & 18.3 & 19.5  & 11.5 & 10.6 & 13.5  & 11.4 & 19.8 & 20.8  & 11.3 & 26.7 & 21.0  & 13.9 & 20.2 & 19.8  & 10.6 & 15.6 & 9.6  & 16.5 \\

Falcon\nocite{yao2025falcon} \pub{arXiv'25} & 14.4 & 6.0 & 9.9 & 1.3 & 0.9 & 3.4 &  9.2 & 5.3 & 8.9 & 3.0 & 1.0 & 0.8 & 3.6 & 1.9 & 2.7 & 12.9 & 4.2 & 7.6   \\

EarthDial\nocite{soni2025earthdial} \pub{CVPR'25} & 14.6 & 18.1  & 10.0 & \underline{27.8} & \underline{25.4} &  \textbf{24.1} & 23.0 & 19.2 & 13.3 & 35.9 & 28.8  & 18.4 & \underline{45.0} & \underline{39.4} &  \textbf{25.9} & 25.8 & 14.8  & 20.4 \\
\midrule
 \textbf{SkyMoE (ours)} &  \textbf{43.0} & \textbf{31.1}  & \underline{19.7} & \textbf{33.4} & \textbf{26.2} &  \underline{19.6} & \underline{37.7} & \textbf{28.3}  & \underline{18.6} & \textbf{60.1} & \textbf{51.3}  & \textbf{38.8} & \textbf{46.1} & \textbf{40.2}  & \underline{25.2} & \textbf{38.7} & \textbf{53.8} &  \underline{37.7}\\
\bottomrule
\end{tabular}
}
\end{center}

\caption{Comparison of SkyMoE with existing generic and RS VLMs on Image Captioning task across multiple benchmarks. B-4, MT, and R-L denote BLUE-4, METEOR, and ROUGE-L scores, respectively.}
\label{tab:ic}
\end{table*}

\begin{table}[!tb]
\begin{center}

	\resizebox{\linewidth}{!}{
	\begin{tabular}{l|ccccccc}
		\toprule
		\multirow{2}{*}{\textbf{Method}} & \multicolumn{3}{c}{\textbf{RSVQA-LR}} & \multicolumn{2}{c}{\textbf{RSVQA-HR}} & \multicolumn{2}{c}{\textbf{MGRS-Bench}} \\
		\cmidrule(lr){2-4} \cmidrule(lr){5-6} \cmidrule(lr){7-8}
		 & \textbf{LR-R} & \textbf{LR-P} & \textbf{LR-C} & \textbf{HR-A} & \textbf{HR-C} & \textbf{MG-P} & \textbf{MG-D}  \\

		\midrule
\multicolumn{7}{l}{\textcolor{gray}{\textit{Close-source Commercial Vision-Language Models}}} \\ 
GPT-4V  \pub{OpenAI'24} & 95.90 & 49.95 & 51.95 & 0 & 68.93 & 92.41 & 42.37    \\ 
\midrule
\multicolumn{8}{l}{\textcolor{gray}{\textit{Open-source Vision-Language Models}}} \\
MiniGPT-v2  & 54.72 & 48.21 & 60.38  & 0  & 59.22 & 58.82 & 40.43 \\
LLaVA-1.5    & 75.47 & 57.14 & 70.59   & 0  & 66.02 & 88.24 & 63.83 \\
Qwen2.5-VL   & 94.34 & 62.50 & 43.14   & 0  & 53.40 & 70.59 & \underline{70.21} \\
DeepSeek-VL  & 62.26 & 48.21 & 64.71  & 0  & 67.96 & 85.29 & 34.04 \\
\midrule 
\multicolumn{8}{l}{\textcolor{gray}{\textit{Open-source Remote Sensing Vision-Language Models}}} \\
GeoChat  & \underline{96.23} & \underline{89.29} & \underline{88.24}  & 23.30 & 75.73 & 94.12 & 63.83 \\
VHM  & 90.57 & 80.36 & 86.27  & 24.27 & \underline{79.61} & 88.24 & 42.55 \\
SkySenseGPT  & 92.45 & 85.71 & 84.31  & 0 & 78.64 & \textbf{95.06} & 63.8 \\
LHRS-Bot  & 84.91 & 76.79 & 49.02  & 0  & 37.86 & 94.32 & 53.19 \\
RSUniVLM & 35.85 & 73.21 & 65.69  & 0 & 76.70 & 85.14 & 27.66 \\
Falcon  & 58.75 & 45.95 & 59.94 & 0 & 42.96 & 68.7 & 42.12 \\
EarthDial & 66.04 & 64.29 & 87.25  & \textbf{34.95} & 71.84 & 76.47 & 46.81 \\
\midrule
		\textbf{SkyMoE (ours)} &\textbf{98.11} & \textbf{91.07} & \textbf{90.20}  & \underline{33.98} & \textbf{80.58} & \underline{94.38}  & \textbf{71.52}\\
		\bottomrule
	\end{tabular}
\vspace{-10pt}	}
\end{center}
\begin{flushleft}
\tiny \textit{\hspace{0.1cm}R/P/C/A/D: Rural / Presence / Comparison / Area / Direction.}
\end{flushleft}
\caption{Comparison of SkyMoE with existing generic and RS VLMs on VQA task. }
\label{tab:vqa}
\end{table}

\begin{table}[!t]
\begin{center}

\resizebox{\linewidth}{!}{
\begin{tabular}{lcccccccccccccccc} 
\toprule
\multirow{2}{*}{\textbf{Method}}&  \multicolumn{2}{c}{\textbf{DIOR-RSVG}} 
& \multicolumn{2}{c}{\textbf{AVVG}} 
& \multicolumn{2}{c}{\textbf{RRSIS-D}} 
& \multicolumn{2}{c}{\textbf{RSVG}} 
& \multicolumn{2}{c}{\textbf{VRSBench}} 
& \multicolumn{2}{c}{\textbf{MGRS-Bench}} \\
\cmidrule(lr){2-3} \cmidrule(lr){4-5} \cmidrule(lr){6-7} \cmidrule(lr){8-9} \cmidrule(lr){10-11} \cmidrule(lr){12-13}
 & \textbf{@0.5} & \textbf{@0.7} 
& \textbf{@0.5} & \textbf{@0.7} 
& \textbf{@0.5} & \textbf{@0.7} 
& \textbf{@0.5} & \textbf{@0.7} 
& \textbf{@0.5} & \textbf{@0.7} 
& \textbf{@0.5} & \textbf{@0.7} \\
\midrule
\multicolumn{7}{l}{\textcolor{gray}{\textit{Close-source Commercial Vision-Language Models}}} \\ 
GPT-4V  \pub{OpenAI'24} & 26.1 & 3.6 & 33.2 & 8.7 &  28.0 & 5.0 & 36.5 & 18.3 & 14.4 & 2.3 & 18.3 & 7.6    \\ 
\midrule
\multicolumn{13}{l}{\textcolor{gray}{\textit{Open-source Vision-Language Models}}} \\
MiniGPT-v2 & 12.2 & 5.3 & 1.5 & 1.0 & 14.0 & 6.5 & 0 & 0 & 12.4 & 5.7 & 3.9 & 0.8 \\
LLaVA-1.5 & 11.4 & 1.6 & 0.5 & 0 & 12.0 & 2.0 & 10.5 & 2.5 & 15.4 & 5.6 & 19.4 & 5.4 \\
Qwen2.5-VL & 36.3 & 19.2 & 0.5 & 0 & 0.5 & 0 & 1.0 & 0 & 45.2 & 25.9 & 1.6 & 0 \\
\midrule 
\multicolumn{13}{l}{\textcolor{gray}{\textit{Open-source Remote Sensing Vision-Language Models}}} \\
GeoChat & 31.4 & 14.7 & 7.5 & 0.5 & 10 & 1.5 & 5.5 & 1 & 56.3 & 32.6 & 17.1 & 10.1 \\
VHM & 55.9 & \underline{42.0} & 0 & 0 & 64.0 & 45.0 & 2.5 & 0.5 & 33.9 & 14.3 & \underline{21.7} & \underline{10.9} \\
SkySenseGPT & \underline{60.8} & 35.5 & \underline{26.5} & \underline{3.5} & \underline{69.0} & 41.5 & 39.5 & 21.5 & \underline{63.5} & \underline{34.2} & 3.1 & 0.7 \\
LHRS-Bot & 7.8 & 2.4 & 0 & 0 & 12.0 & 1.5 & 1.0 & 0 & 1.0 & 0.2 & 1.6 & 0 \\
GeoGround  & 37.6 & 22.0 & 20.0 & 9.0 & 59.5 & 39.5 & 22.5 & 11.0 & 41.4 & 21.6 & 7.0 & 2.3 \\
EarthDial  & 46.1 & 34.3 & 3.5 & 2.0 & \textbf{72.5} & \underline{55.5} & \underline{42.0} & \underline{24.0} & 14.4 & 9.7 & 17.1 & 9.3 \\
\midrule
\textbf{SkyMoE (Ours)}&  \textbf{68.6} & \textbf{48.6} & \textbf{36.5} & \textbf{18.0} & \textbf{72.5} & \textbf{56.1} & \textbf{42.5} & \textbf{26.0} & \textbf{66.1} & \textbf{44.4} & \textbf{22.3} & \textbf{13.4} \\
\bottomrule
\end{tabular}

}
\end{center}
\caption{Comparison of SkyMoE with generic and RS VLMs on Visual Grounding task using mAP@0.5 and @0.7.}
\label{tab:vg}
\end{table}

\subsection{Context-Disentangled Data Augmentation}

To enhance the diversity and granularity sensitivity of the training data, we design augmentation strategies that explicitly vary object count, position, and appearance—encouraging SkyMoE to specialize across experts and better capture multi-scale semantics. 

\textbf{Count-Varying Cutout.}  
To simulate diverse object densities for counting tasks, we randomly mask a portion of objects in each image. Given an original count $N_c$, we sample a masking ratio $r \sim \mathcal{U}(0.15, 0.30)$ and remove $m = \lceil r \cdot N_c \rceil$ objects by zero-filling their regions. Updated annotations reflect the new count $N_{\text{new}} = N_c - m$. Each image yields four variants with at least $\lceil 0.1N_c \rceil$ difference in count, enabling fine-grained supervision for density-sensitive predictions.

\textbf{Attribute Editing.}  
To strengthen vision-language alignment, we manipulate referring expressions by editing spatial and color attributes, then apply corresponding visual transformations. For spatial editing, positional phrases (e.g., “bottom”) are replaced (e.g., “top right”) and the target object is relocated within a $3\times3$ grid cell, ensuring visual realism via Poisson blending and minimal overlap ($\text{IoU} < 0.1$). For color editing, textual color terms are swapped (e.g., “green” $\rightarrow$ “red”), and matched via palette transfer while preserving material texture and reflectance.

These granularity-aware augmentations diversify scene configurations while maintaining annotation consistency, allowing SkyMoE’s experts to specialize in distinct visual patterns and improving overall generalization.

\subsection{Multi-Granularity RS Benchmark} 

To address the limitations of existing RS-VLM benchmarks, such as limited task diversity, insufficient granularity control, and restricted resolution variation, we introduce MGRS-Bench, a dedicated evaluation suite for RS-VLMs.

MGRS-Bench is constructed through a multi-stage annotation pipeline consisting of four components: (1) attribute extraction to identify relevant visual features and contextual cues, (2) prompt engineering to generate diverse instruction formats, (3) GPT-4 inference to produce high-quality candidate responses, and (4) human verification to ensure accuracy, disambiguation, and label consistency. The final dataset includes 10,415 RS images and 18,433 annotated instances, with 2,083 images assigned to each of the five tasks.

\section{Experiments}
\subsection{Evaluation Results}

To thoroughly assess the performance of various VLMs in RS tasks, we employ a comprehensive evaluation scheme that includes Image Captioning (IC), Visual Question Answering (VQA), Visual Grounding (VG), Object Counting (OC), and Scene Classification (SC). This evaluation scheme is designed to provide a holistic view of each model's capabilities across a spectrum of tasks, from high-level reasoning to detailed perceptual tasks.

\textbf{Image Captioning.} 
SkyMoE achieves competitive performance in IC, demonstrating robust generalization across six public benchmarks~\cite{lu2017exploring,yuan2021exploring,chen2022parallel,qu2016deep}. As shown in ~\Cref{tab:ic}, it obtains the highest BLEU-4 scores on UCM-Captions (43.0), RSICD (33.4), and NWPU-Captions (60.1), and leads in METEOR on six datasets, including a strong 53.8 on MGRS-Bench. These results highlight SkyMoE’s superior ability to generate accurate, fluent, and context-aware descriptions. Compared to existing RS-VLMs, most of which can recognize object pairs or isolated features, SkyMoE demonstrates a clear advantage in modeling complex spatial and semantic relations. For example, while SkySenseGPT performs competitively in grounding tasks, its performance on RSICD and MGRS-Bench captioning drops significantly (e.g., BLEU-4: 12.2 and 24.9), suggesting a limited capacity to structure local features into coherent narratives. In contrast, SkyMoE’s consistent gains across BLEU\nocite{papineni2002bleu}, METEOR\nocite{banerjee2005meteor}, and ROUGE-L\nocite{lin2004rouge} indicate its holistic reasoning capability, empowered by MoE-based expert specialization and fine-grained data augmentation.

\textbf{Visual Question Answering.} 
As shown in \Cref{tab:vqa}, SkyMoE consistently outperforms all baselines across diverse VQA scenarios~\cite{lobry2020rsvqa}. It excels on both low-resolution tasks that require global contextual understanding (e.g., 98.11\% on rural classification) and high-resolution tasks that demand fine-grained spatial precision (e.g., 33.98\% on area estimation). Notably, SkyMoE maintains strong performance across resolution levels, as seen in the comparison tasks (90.20\% on LR-C vs. 80.58\% on HR-C), indicating its robust ability to balance local and global feature representations. The resolution gap in RS-VQA fundamentally tests a model's capacity to generalize across different perceptual scales: low-resolution questions rely on coarse semantic cues, whereas high-resolution ones require precise detail recognition. SkyMoE effectively bridges this gap through the synergy between its MoE architecture and context-disentangled augmentation strategy. This dynamic adaptation mechanism allows the model to tailor its representation to task-specific granularity, setting a new benchmark for resolution-variant remote sensing VQA.

\begin{table*}[!htb]
\begin{center}

\resizebox{\linewidth}{!}{
\begin{tabular}{l|cccc|cccccccc}
\toprule
\multirow{2}{*}{\textbf{Dataset}} & \multicolumn{4}{c|}{\textbf{Open-source Vision-Language Models}} & \multicolumn{8}{c}{\textbf{Open-source Remote Sensing Vision-Language Models}} \\
\cmidrule(lr){2-5} \cmidrule(lr){6-13}
& MiniGPT-v2 & LLaVA-1.5 & Qwen2.5-VL & DeepSeek-VL & GeoChat & VHM & SkySenseGPT & LHRS-Bot & RSUniVLM & EarthDial & Falcon & \textbf{SkyMoE} \\
\midrule
RESISC45 & 21.78 & 42.22 & 69.78 & 46.00 & 84.67 & \underline{91.33} & 83.33 & 42.44 & 2.22 & 76.67 & 62.22 & \textbf{91.77} \\
AID  & 18.83 & 47.33 & 63.83 & 34.67 & 71.50 & \textbf{79.00} & 75.50 & 39.00 & 3.33 & 62.50 & 20.83 & \underline{78.21} \\
WHU-RS19 & 26.84 & 62.11 & 79.47 & 80.00 & 89.47 & 91.84 & \textbf{93.16} & 56.58 & 5.26 & 73.42 & 44.21 & \underline{92.37} \\
MGRS-Bench & 35.10 & 59.35 & \textbf{76.44} & 44.80 & 56.35 & 66.28 & 55.89 & 51.04 & 7.62 & 42.96 & 38.89 &  \underline{68.73} \\
\bottomrule
\end{tabular}
}

\end{center}
\caption{Comparison of SkyMoE for Scene Classification tasks across multiple benchmarks.}
\label{tab:sc}
\end{table*}

\begin{table}[!t]
\begin{center}

\resizebox{\linewidth}{!}{
\begin{tabular}{l|cccccc} 
\toprule
\textbf{Method} & \textbf{RSOD} & \textbf{HRRSD} & \textbf{NWPU-VHR} & \textbf{DOTAv2} & \textbf{VisDrone} & \textbf{MGRS} \\
\midrule
\multicolumn{7}{l}{\textcolor{gray}{\textit{Close-source Commercial Vision-Language Models}}} \\ 
GPT-4V  \pub{OpenAI'24} & 40.0 & 58.5 & 58.0 & 19.7 & 19.0 &  28.4     \\ 
\midrule
\multicolumn{7}{l}{\textcolor{gray}{\textit{Open-source Vision-Language Models}}} \\
MiniGPT-v2 &20.0 & 26.5&23.0&7.7 &8.5 & 4.6 \\
LLaVA-1.5 &42.5&52.4&40.5&11.8 &10.5& 21.5  \\
Qwen2.5-VL  &40.5 &56.7 &57.5 &17.4 &14.0 & 27.7  \\
DeepSeek-VL  &\underline{61.0} &\textbf{61.5}&\underline{60.0} & 18.3 &\underline{20.5} & 29.2  \\
\midrule 
\multicolumn{7}{l}{\textcolor{gray}{\textit{Open-source Remote Sensing Vision-Language Models}}} \\
GeoChat  &19.5 &57.6&42.5 & 16.9  & 14.5 & 21.5 \\
VHM   &16.0 & 46.7 &48.5 & 18.0 & 5.5 & \underline{30.8} \\
SkySenseGPT  &51.5 & \underline{58.7} &49.5 &\underline{21.2} &18.5 & 27.7  \\
LHRS-Bot  &2.0 &2.3 &3.0 & 10.5 &1.5 & 12.3  \\

RSUniVLM  & 49.5 & 54.2 & 36.0 & 19.0 & 18.5 & 27.5 \\
Falcon  & 47.0 & 42.6 & 52.5 & 18.1 & 3.5 & 26.4 \\
EarthDial &41.0&61.5&52.5&20.9&18.0 & 29.2 \\
\midrule
\textbf{SkyMoE (ours)} &\textbf{67.6} &57.8 &\textbf{62.3} & \textbf{26.4} &\textbf{21.5} & \textbf{31.6} \\
\bottomrule
\end{tabular}

}
\end{center}
\caption{Comparison of SkyMoE for Object Counting tasks across various datasets.}
\label{tab:oc}
\end{table}

\textbf{Visual Grounding.} 
SkyMoE delivers consistent and competitive performance across all VG benchmarks~\cite{zhan2023rsvg,zhou2024geoground,liu2024rotated,sun2022visual,NEURIPS2024_05b7f821}. As shown in ~\Cref{tab:vg}, it achieves a strong mAP@0.5 of 68.6\% on DIOR-RSVG, with only moderate degradation at mAP@0.7 (48.6\%), indicating its robust ability to balance global scene comprehension with precise object localization. In contrast, generic VLMs like MiniGPT-v2 and LLaVA-1.5 struggle to surpass even 12\% at mAP@0.5 and collapse almost entirely at stricter thresholds (e.g., MiniGPT-v2: 12.2\% → 5.3\%), reflecting over-reliance on global semantics while lacking spatial precision. RS-VLMs (e.g., GeoChat, VHM, SkySenseGPT) exhibit stronger low-threshold performance (e.g., SkySenseGPT: 60.8\% mAP@0.5 on DIOR-RSVG), but their performance degrades significantly at mAP@0.7 (e.g., SkySenseGPT: 35.5\%), revealing a deficiency in handling fine-grained grounding. On AVVG, which features high spatial ambiguity, SkyMoE further distinguishes itself by maintaining 36.5\% (mAP@0.5) and 18.0\% (mAP@0.7), while most baselines fall below 5\%. These results underscore SkyMoE’s robustness in integrating global and local cues, a key capability for real-world RS applications.

\textbf{Object Counting.} 
As shown in ~\Cref{tab:oc}, SkyMoE demonstrates strong density-invariant performance, excelling in both sparse (e.g., 67.6\% on RSOD) and ultra-dense (e.g., 26.4\% on DOTAv2) aerial scenes~\cite{long2017accurate,zhang2019hierarchical,cheng2014multi,xia2018dota,zhu2021detection}, showcasing a remarkable ability to dynamically balance fine-grained local details and global contextual cues. In sparse environments, it precisely detects individual instances, while in dense scenes, it preserves instance awareness without losing structural coherence. This robustness stems from the synergy of our MoE architecture and the proposed count-varying cutout augmentation, which explicitly\begin{table}[!htb]
\begin{center}

\resizebox{\linewidth}{!}{
\begin{tabular}{l|ccccc}
\toprule
\multirow{2}{*}{\textbf{Method}} & \textbf{VG} & \textbf{OC} & \textbf{IC} & \textbf{VQA} & \textbf{SC} \\
                & (mAP@0.5) & (Acc) & (BLEU-4) & (Acc) & (Acc) \\
\midrule
RSUniVLM & 17.1 & 27.7 & 15.6 & 56.4 & 7.6 \\
MoE-LLaVA & 18.9 & 28.2 & 20.7 & 61.0 & 11.6 \\
\midrule
\textbf{SkyMoE (ours)} & \textbf{22.3} & \textbf{31.6} & \textbf{38.7} & \textbf{83.0} & \textbf{68.7} \\
\bottomrule
\end{tabular}
}

\end{center}
\caption{Comparison with other MoE-based VLMs.}
\label{tab:vl_tasks}
\end{table}

\begin{table}[!htb]
\begin{center}

\resizebox{\linewidth}{!}{
\begin{tabular}{l|cccc|c}
\toprule
\textbf{Metric} & \textbf{GeoChat} & \textbf{VHM} & \textbf{SkySenseGPT} & \textbf{LHRS-Bot} & \textbf{Ours} \\
\midrule
Trainable Parameters  & 7.06B & 6.77B & 7.05B & 6.74B & 9.36B \\
TFLOPs                & 694.31 & 665.30 & 694.31 & 662.41 & 712.56 \\
\midrule
Per-token Latency (ms) & 26.92 & 21.92 & 24.59 & \textbf{18.97} &  \underline{20.83} \\
\bottomrule
\end{tabular}
}

\end{center}
\caption{Results on training and inference efficiency.}
\label{tab:efficiency}
\end{table}
\noindent trains the model to adapt to density shifts. SkyMoE also achieves a new state-of-the-art on NWPU-VHR (62.3\%), further establishing its superiority in the challenging task of aerial object counting.

\textbf{Scene Classification.} 
SkyMoE achieves highly competitive results on scene classification benchmarks~\cite{cheng2017remote,xia2017aid,xia2010structural} (e.g., 78.21\% on AID, 92.37\% on WHU-RS19), all within 2\% of the top-performing models. This small gap reflects a deliberate design trade-off: unlike models that solely optimize for global semantics, SkyMoE’s MoE framework allocates capacity to preserve fine-grained local features. While slightly limiting performance on purely global tasks, it enables SkyMoE to outperform all baselines on more demanding tasks like visual grounding and object counting, resulting in a more robust and versatile model for diverse geospatial challenges.

\textbf{Comparison with MoE-based VLMs.}
As shown in Table~\ref{tab:vl_tasks}, SkyMoE achieves the best performance across all five RS vision-language tasks on MGRS-Bench, significantly outperforming previous MoE-based models such as RSUniVLM and MoE-LLaVA. SkyMoE achieves over +18 BLEU-4 improvement in IC. The substantial gains in high-level reasoning tasks like VQA (+22.0 Acc) and SC (+57.1 Acc) further demonstrate the model’s strong capability in both fine-grained perception and global semantic understanding. 

\textbf{Training and Inference Efficiency.}
We compare the training and inference costs of SkyMoE with four recent RS-VLMs in Table~\ref{tab:efficiency}. While SkyMoE has a larger number of trainable parameters, its overall training complexity remains comparable to other models. It also achieves competitive per-token latency, outperforming larger models such as GeoChat and SkySenseGPT. This efficiency is attributed to the top-k expert activation mechanism, which preserves model capacity while reducing inference overhead. Compared with LHRS-Bot's fully shared architecture, SkyMoE offers a better trade-off between scalability and efficiency.

\begin{table}[!tb]
\begin{center}

\resizebox{\columnwidth}{!}{
\begin{tabular}{ccc|ccccc}
\toprule
 \multirow{2}{*}{\textbf{\#Expert}} 
&\multicolumn{2}{c|}{\textbf{CDA}} & \textbf{VG} & \textbf{IC} & \textbf{OC} & \textbf{VQA} & \textbf{SC} \\
\cmidrule(lr){2-3}
 & \textbf{CVC} & \textbf{AE} & \textbf{DIOR-RSVG} & \textbf{UCM-Cap} & \textbf{RSOD} & \textbf{RSVQA} & \textbf{RESISC45} \\
\midrule
\multicolumn{8}{l}{\textit{Varying number of experts (with both modules enabled)}} \\
1  &\checkmark & \checkmark &  45.40 & 45.4 & 44.50 & 74.67 & 69.67 \\
4  &\checkmark & \checkmark &  63.79 & 67.4 & 58.49 & 83.64 & 80.20 \\
6  &\checkmark & \checkmark &  67.64 & 73.2 & 66.65 & 91.51 & 89.26 \\
8  &\checkmark & \checkmark &  \textbf{68.60} & \textbf{78.8} & \textbf{68.51} & \textbf{93.13} & \textbf{91.77} \\
\midrule
\multicolumn{8}{l}{\textit{Ablating key components (fixed \#Expert = 8)}} \\
8 &\xmark     & \xmark      & 60.20 & 72.0 & 56.37 & 82.54 & 91.29 \\
8 &\checkmark & \xmark      & 60.34 & 72.1 & 65.84 & 89.88 & \textbf{91.86} \\
8 &\xmark     & \checkmark  & 68.10 & 77.6 & 61.23 & 86.57 & 90.11 \\
 8 &\checkmark & \checkmark  & \textbf{68.60} & \textbf{78.8} & \textbf{68.51} & \textbf{93.13} & 91.77 \\
\bottomrule
\end{tabular}
}
\end{center}
\begin{flushleft}
\fontsize{5.5pt}{6pt} \textit{\hspace{0.1cm}CDA: Context-Disentagled Augumentation, CVC: Count-Varying Cutout, AE: Attribute Editing}
\end{flushleft}
\caption{Performance with different architectures.}
\label{tab:ablation_simplified}
\end{table}

\textbf{Ablation study.} Our ablation study dissects SkyMoE to validate the contribution of its components. First, we demonstrate the efficacy of the MoE architecture itself, as performance scales positively with the number of experts (e.g., VG improves from 45.4\% to 68.6\% as experts increase from 1 to 8). Second, our purpose-built data augmentation strategy provides a significant performance uplift, proving particularly effective for tasks sensitive to local details such as object counting (+9.47\%). Critically, the full framework significantly outperforms any partial configuration. For example, the complete model achieves a VQA score of 93.13\%, a +6.56 gain over using the MoE architecture with standard training. This non-additive performance gain proves that our SOTA results are not achieved through isolated improvements, but through a co-designed framework where our targeted data augmentation actively unlocks and directs the latent potential of the MoE architecture.

\section{Conclusion}

In this work, we proposed SkyMoE, a novel framework that combines a MoE architecture with a tailored data augmentation strategy to promote expert specialization. Unlike prior methods, SkyMoE dynamically allocates experts based on feature granularity, enabling a robust balance between local and global representations. To support systematic evaluation, we constructed MGRS-Bench, a comprehensive benchmark covering diverse RS tasks and granularity levels. Extensive experiments on 21 datasets demonstrate SOTA performance, with ablations and visualizations confirming that this improvement stems from our model’s learned ability to adaptively route information. 

\section*{Acknowledgments}

This work was supported by the National Key Research and Development Program of China (Grant No. 2021ZD0112500), National Natural Science Foundation of China (Grant No. U22A2098, 62172185, 62206105 and 62202200), Major Science and Technology Project of Jilin Province (Grant No. 20240212001GX), and Major Science and Technology Project of Changchun City (Grant No. 2024WX05).

\bibliography{aaai2026}

\end{document}